\documentclass[runningheads]{llncs}
\usepackage[T1]{fontenc}
\usepackage{booktabs}
\usepackage{multirow} 
\usepackage{graphicx} 
\usepackage{generic}
\usepackage{cite}
\usepackage{amsmath,amssymb,amsfonts}
\usepackage{algorithmic}
\usepackage{textcomp}
\usepackage[table,xcdraw]{xcolor}
\usepackage{colortbl}
\usepackage{float}

\begin{document}
\title{UlRe-NeRF: 3D Ultrasound Imaging through Neural Rendering with Ultrasound Reflection Direction Parameterization}
%
%
\author{Ziwen Guo\and
Zi Fang\and
Zhuang Fu*}
\authorrunning{Z. Guo et al.}
\institute{Shanghai Jiao Tong University \\
\email{miixnon1@sjtu.edu.cn}
\vspace{\baselineskip}
}
%

%
\maketitle              
\begin{abstract}

Three-dimensional ultrasound imaging is a critical technology widely used in medical diagnostics. However, traditional 3D ultrasound imaging methods have limitations such as fixed resolution, low storage efficiency, and insufficient contextual connectivity, leading to poor performance in handling complex artifacts and reflection characteristics. Recently, techniques based on NeRF (Neural Radiance Fields) have made significant progress in view synthesis and 3D reconstruction, but there remains a research gap in high-quality ultrasound imaging. To address these issues, we propose a new model, UlRe-NeRF, which combines implicit neural networks and explicit ultrasound volume rendering into an ultrasound neural rendering architecture. This model incorporates reflection direction parameterization and harmonic encoding, using a directional MLP module to generate view-dependent high-frequency reflection intensity estimates, and a spatial MLP module to produce the medium's physical property parameters. These parameters are used in the volume rendering process to accurately reproduce the propagation and reflection behavior of ultrasound waves in the medium. Experimental results demonstrate that the UlRe-NeRF model significantly enhances the realism and accuracy of high-fidelity ultrasound image reconstruction, especially in handling complex medium structures.

\end{abstract}
%
%
%
\section{Introduction}
\label{sec:introduction}
Ultrasound (US) is a highly valued imaging modality in medicine due to its cost-effectiveness, safety, non-invasiveness, and ability to provide real-time imaging. Unlike MRI and CT, US is portable, radiation-free, and user-friendly, making it ideal for various clinical applications, including intraoperative imaging and surgical guidance. Its affordability and efficiency have led to its growing popularity among clinicians and researchers. Compared to conventional two-dimensional (2D) ultrasound, three-dimensional (3D) ultrasound offers significant advantages. 3D ultrasound allows direct visualization of 3D anatomy, enhancing diagnostic accuracy and reducing dependence on operator skill, while providing more precise measurements of organ volumes and other quantitative properties, enabling repeatable and consistent imaging for follow-up studies\cite{fenster2001three}. It also offers the flexibility to view anatomical structures from any orientation, overcoming the limitations imposed by patient positioning in 2D ultrasound. 

Three-dimensional (3D) ultrasound imaging involves several key steps to realize view synthesis of novel 2D ultrasound images\cite{hsu2009freehand}\cite{mercier2005review}. The first step is the data acquisition phase, where a series of 2D ultrasound images $I_i \in \mathcal{J}$ along with corresponding image plane pose $T_{I_i} \in \mathcal{T}$ are captured using a 1D ultrasound probe with tracking sensors attached. The next step is the volumetric representation phase, where a 3D representation $\hat{V}=F_\theta(I, T)$ is obtained based on the 2D images and their corresponding poses, aiming to preserve as much of the information as possible. The final step is the rendering phase, for any novel image plane pose $T_{l_{\text {low }}} \in \mathcal{T}$, a synthesized 2D ultrasound image $\hat{I}_{\text {new }}=G_\mu\left(\hat{V}, T_{\text {new }}\right)$ can be rendered. A high-quality 3D scene reconstruction should achieve synthesized views that are as close to the actual views as possible. 

Much past decades, some researches about explicit volumetric representation have recently been developed\cite{mercier2005review}. Each slice is treated as a slice point cloud $P_i \in \mathcal{P}$, where each point has corresponding spatial coordinates $x_i$ and color $c_i$. In the volumetric representation phase, an explicit 3D voxel grid $\hat{V}$ is constructed. For each voxel $x_{\text {voxel }} \in \hat{V}$, the corresponding color $\hat{c}\left(x_{\text {voxel }}\right)$ is obtained by using a parameterized interpolation algorithm $F_\theta(\{\cdot\})$ to gather the spatial and color information of the point cloud points contained in its spatial neighborhood $N_\delta\left(x_{\text {voxel }}\right) \cap \mathcal{P}$.

However, those explicit volumetric representation methods have some flaws, leading to that synthetic images do not conform to physical laws:

\begin{figure*}[!t]
\centerline{\includegraphics[width=\columnwidth]{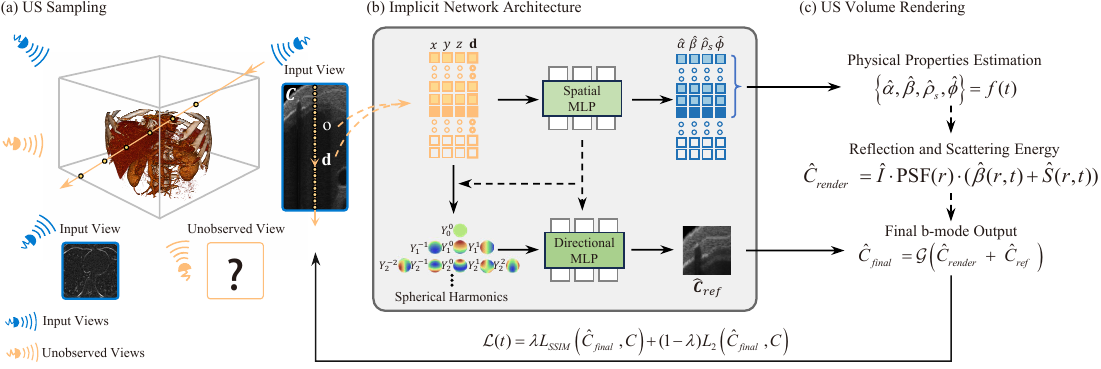}}
\caption{UlRe-NeRF for High-Fidelity 3D Ultrasound Imaging. This figure illustrates our proposed ultrasound neural rendering architecture, which integrates implicit neural representation (b)\label{fig1b} and volume rendering (c)\label{fig1c} to enhance 3D ultrasound imaging. The system comprises two main modules: the spatial MLP and the directional MLP. The spatial MLP predicts physical property parameters, such as attenuation coefficients, reflectivity, scattering density, and amplitude, through high-dimensional position encoding. This simulates the propagation and reflection behavior of ultrasound waves within the medium. The directional MLP processes the input ray directions, generating view-dependent intensity estimates that capture complex reflection phenomena.}
\label{fig1}
\end{figure*}

\begin{enumerate}
\item The unreasonable Lambertian assumption, that the same spatial point appears the same from different viewpoints, causes difficulty in reproducing physical phenomena about viewpoints like refraction and mirror image artifacts.
\item Spatial neighborhood interpolation methods have a very small receptive field, which weaken the contextual connections between spatially distant points, making it hard to reproduce acoustic shadowing and enhancement caused by different tissue impedances.
\item Explicit discrete voxel grids are inefficient in terms of storage and have a fixed resolution, which depends on the grid and is often very low. Querying the values between grid points requires linear interpolation. Furthermore, the discretization makes differentiation difficult.
\end{enumerate}

In the literature, as summarized in Section 2.1, researchers attempt to address this issue through implicit representation methods. In explicit representation, the color value of a query point is obtained by searching the spatial neighborhood in the dataset and interpolating $F_\theta:\left\{\left(N_\delta(x) \cap \mathcal{P}\right)\right\} \rightarrow \hat{c}$. In implicit representation, uses a continuous coordinate-based MLP (Multi-Layer Perceptron) $F_\theta:(x, d) \rightarrow \hat{f}$ to query multidimensional representations $\hat{f}$ from three-dimensional coordinates $x$ and two-dimensional directions $d$.

Neural Radiance Fields (NeRF) have significantly advanced the general application of implicit representation\cite{tewari2022advances}. NeRF uses implicit representation, specifically a coordinate-based MLP supervised to represent shape or appearance, combined with explicit rendering through a fixed differentiable rendering engine, to achieve novel view synthesis. The strong physically motivated inductive biases act as regularizers, ensuring that the derived implicit representation closely mimics the real world, which leads to better generalization. 

Based on the foundational concepts of NeRF and incorporating additional inductive biases, numerous NeRF variants have been proposed to expand their application scenarios, such as ultrasound, CT, and MRI, or to enhance performance, such as improving imaging quality and enabling sparse data training. 

Ultra-NeRF is the first work to bring NeRF into the field of ultrasound\cite{wysocki2024ultra}. It builds on the ultrasound volume rendering method by Salehi et al. and combines a coordinate-based MLP that maps from 3D coordinates to five physical parameters, achieving high-quality synthetic ultrasound images\cite{salehi2015patient}. However, it is still view-independent, so cannot achieve ray interactions and the Fresnel Effect, and it does not allow rendering complex ultrasound artifacts, such as reverberations. To address these issues, we drew inspiration from Ref-NeRF and proposed a new model that considers both reflection intensity and volume rendering (Fig. \ref{fig1b}(b) and \ref{fig1c}(c)). Specifically, we present a ultrasound neural rendering architecture based on implicit neural networks and volume rendering, optimizing the rendering process to better simulate the physical properties, artifacts, and deformations of the medium. We call this model UlRe-NeRF, which can accurately reproduce the propagation and reflection behavior of ultrasound waves within the medium, as depicted in Fig. \ref{fig1}.
In summary, our main contributions are as follows:

\subsubsection{High-fidelity 3D Ultrasound Imaging}
{We introduce a novel model that synthesizes precise ultrasound images by learning view-dependent appearance and geometry from multiple ultrasound scans. This method significantly enhances image quality, particularly in capturing fine details and complex tissue structures. It not only improves diagnostic accuracy but also reduces reliance on operator expertise.}
\subsubsection{Enhanced Implicit Representation Network}
{We designed an implicit representation network that more accurately models the propagation characteristics of ultrasound waves. By incorporating advanced techniques such as reflection direction parameterization and Reflective Harmonic Encoding (RHE), the network better captures high-frequency information and complex tissue reflections. These improvements enable our method to more precisely reconstruct ultrasound images when dealing with intricate tissue structures.}
\subsubsection{Ultrasound Volume Rendering}
{Building on the enhanced implicit representation network, we developed a complex rendering model that integrates reflection intensity and volume rendering. This model accurately simulates the medium's physical properties, artifacts, and deformations, and correctly handles light interactions and Fresnel effects within the medium. This advancement is crucial for achieving realistic ultrasound image synthesis, effectively overcoming the limitations of existing methods in simulating complex physical phenomena.}

In conclusion, UlRe-NeRF has made significant strides in the field of 3D ultrasound imaging, excelling in both image quality and physical realism. It also shows broad potential for applications in medical imaging. Our research represents an important advancement in ultrasound image reconstruction technology and lays a solid foundation for future research and applications.

\section{Related Works}
\subsection{Variants of NeRF for quality}
Neural Radiance Fields (NeRF) uses implicit representation and explicit rendering to achieve novel view synthesis\cite{mildenhall2021nerf}. Given any input 3D position $x$ and view direction $d$, NeRF uses a spatial MLP to output the density $\tau(x)$ and bottleneck vector $b(x)$ from 3D position $x$, and then use a second directional MLP to outputs the color of light $c(x, d)$  emitted by a particle at 3D position $x$ and direction $d$ (see Fig. \ref{fig3} for a visualization). The densities $\left\{\tau_i(x)\right\}$ and colors $\left\{c_i(x, d)\right\}$ queried along the camera ray $x_i=o+t_i d, d_i=d$ originating at $o$ with direction $d$, are alpha composited \cite{max1995optical} to obtain the color of the pixel corresponding to the ray. 
\begin{equation}
\begin{aligned}
\hat{C}(o, d) &=\sum_i w_i c_i, w_i \\
&=e^{-\sum_{j<i} \tau_i\left(t_{j+1}-t_j\right)}\left(1-e^{-\tau_i\left(t_{j+1}-t_j\right)}\right)
\end{aligned}
\end{equation}

Numerous subsequent NeRF variants focus on improving the core component of NeRF to improve imaging quality. MLP tends to learn low-frequency functions, limiting its ability to represent complex geometry and texture \cite{rahaman2019spectral}. To address this issue, NeRF maps the Cartesian coordinates to a higher dimensional space using sinusoidal positional encoding \cite{mildenhall2021nerf} \cite{tancik2020fourier}. Mip-NeRF \cite{barron2021mip} improved volume rendering by using cone tracing instead of the ray tracing in standard NeRF. They introduced Integrated Positional Encoding (IPE), where a cone cast from the camera’s center through the pixel was approximated by a multivariate Gaussian. By IPE over a conical frustum for each quadrature segment along the ray, Mip-NeRF encoded the scene at multiple scales, preventing aliasing when rendering from different positions or resolutions. However, due to differences in imaging principles, these neural volumetric scene representation methods designed for natural images are not well-suited for reconstructing the complex biological tissues in ultrasound images. Consequently, their view synthesis performance falls short of meeting the requirements for 3D ultrasound reconstruction.

Ref-NeRF \cite{verbin2022ref} enhances NeRF by structuring view-dependent appearance into pre-filtered incident radiance, diffuse color, material roughness, and specular tint. This decomposition method allows for more accurate rendering of specular highlights and reflections. To achieve this, view direction reflections are reparameterized as a function of reflection direction, significantly improving the interpolation of reflective appearances. The specific formula is as follows:
\begin{equation}
\mathbf{r}=2(\mathbf{v} \cdot \mathbf{n}) \mathbf{n}-\mathbf{v}
\end{equation}
Additionally, Ref-NeRF introduces Integrated Directional Encoding (IDE) to effectively represent the material radiance function as it varies with roughness. IDE encodes the reflection direction using a convolution of spherical harmonics and von Mises-Fisher distribution:
\begin{equation}
\operatorname{IDE}(\mathbf{r})=\sum_{l=0}^L \sum_{m=-l}^l A_{l m} \cdot Y_{l m}(\mathbf{r})
\end{equation}

This method captures finer high-frequency information in reflection directions and shares lighting information across different roughness regions, significantly enhancing rendering realism and accuracy. By modeling reflection directions based on computer graphics principles, it provides a viable solution for accurately capturing ultrasound wave reflections and scattering at tissue interfaces, thereby improving image detail and clarity.

In summary, although NeRF may not be directly suitable for 3D ultrasound reconstruction due to differences in data characteristics and imaging principles, we can leverage the capability of implicit neural representations to reconstruct high-quality images from sparse viewpoints. By combining volume rendering, we can develop new models tailored for ultrasound imaging, addressing the reflections and refractions of ultrasound waves at various tissue interfaces, thus generating more realistic ultrasound images.

\subsection{Implicit neural representations in medical imaging}
Implicit neural models have been widely used in medical imaging, particularly in solving inverse imaging problems\cite{huang2017review}. These problems involve learning the structure of an object (such as an organ of interest in medical cases) from observations or measurements. By using INRs, it becomes possible to synthesize novel high-resolution scans from sparsely-sampled low-resolution data with minimal memory requirements.

Some research has been conducted on reconstructing various medical images using innovative approaches.For MRI modality, IREM \cite{wu2021irem} and ArSSR \cite{wu2022arbitrary} leverage implicit continuous representations to recover high-resolution MRI images from low-resolution ones, achieving significant improvements in anatomical detail and image contrast.For CT modality, NeRP \cite{shen2022nerp} reconstructs CT images from sparsely sampled measurements and incorporates longitudinal prior knowledge into deep learning frameworks as network parameters.For X-ray modality, Med-NeRF \cite{corona2022mednerf} explores the inference of anatomical 3D structures from a few or a single-view X-ray by combining GRAF \cite{schwarz2020graf} with discriminator architecture adapted for CT-projections. In the field of ultrasound, ImplicitVol \cite{yeung2021implicitvol} and work by Li et al. \cite{li20213d} are the early explorations utilizing implicit representations and neutral rendering for high-quality 3D ultrasound reconstruction. But they pay little attention to the unique features of ultrasound, lacking some strong physically motivated inductive biases as regularizers. This results in poor image quality and defects that violate physical laws.

\begin{figure*}[!t]
\centerline{\includegraphics[width=\columnwidth]{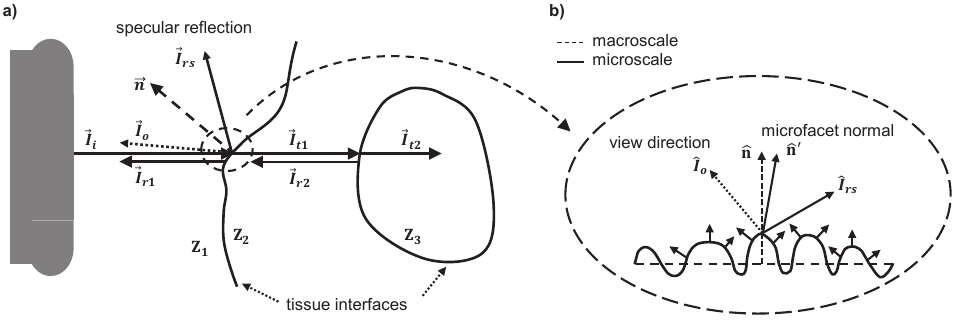}}
\caption{Ultrasound Wave Interaction and Reflection Parameterization at the Medium Interface. (a)\label{fig2a} Shows the reflection and refraction phenomena when the ultrasound wave $\boldsymbol{I}_i$ impinges on the medium interface, describing the basic physical processes and corresponding formulas. (b)\label{fig2b} Describes the process of calculating the specular reflection direction $\hat{\boldsymbol{I}}_{r s}$ by adjusting the macroscale normal vector $\widehat{\mathbf{n}}$ to the microfacet normal vector $\widehat{\mathbf{n}}^{\prime}$, highlighting the impact of the microfacet structure on reflection characteristics.}
\label{fig2}
\end{figure*}

To address these issues, Ultra-NeRF \cite{wysocki2024ultra} propose an implicit neural representation with physics-based rendering for US imaging. It builds on the ultrasound volume rendering method by Salehi et al. \cite{salehi2015patient} and combines a coordinate-based MLP that maps from 3D coordinates to five physical parameters, achieving high-quality synthetic ultrasound images. However, this approach is still view-independent, preventing it from achieving ray interactions and the Fresnel Effect. It also does not allow for rendering complex ultrasound artifacts, such as reverberations. Our work draws on this combination of traditional ray tracing and physics-based volume rendering techniques, achieving better reconstruction results for 3D ultrasound scenes through implicit neural methods.

\section{Methods}
NeRF and its early variants perform poorly in parameterizing reflection directions and handling high-frequency information, especially in complex media structures and high-fidelity image rendering\cite{debbagh2023neural}. These models typically fail to accurately capture complex reflection phenomena, resulting in suboptimal rendering quality. Ultra-NeRF, the first to introduce NeRF concepts into ultrasound imaging, also faces limitations in handling light interactions and Fresnel effects, further affecting its performance in rendering complex artifacts and reflection characteristics\cite{wysocki2024ultra}. To address these issues, we proposed a new model, UlRe-NeRF, which considers both reflection intensity and ultrasound volume rendering. Specifically, we designed an ultrasound neural rendering architecture based on implicit neural networks and volume rendering, optimizing the rendering process (Figure 3) to better simulate the medium's physical characteristics, artifacts, and deformations, accurately reproducing the propagation and reflection behavior of ultrasound waves within the medium. Our approach provides a more effective and precise solution for ultrasound imaging.

\subsection{Ultrasound Reflection Direction Parameterization}

To better simulate the reflection characteristics of ultrasound waves at the medium interface, we propose a new method for parameterizing reflection directions, as shown in Figure 2(a). The reflection and refraction properties of ultrasound waves, treated as rays, can be calculated using the laws of wave physics\cite{shams2008real}. When an ultrasound wave $\boldsymbol{I}_i$ impinges on the medium interface at a certain angle, reflection and refraction occur. The intensities of reflection and refraction can be calculated using the following formulas:
\begin{equation}
I_r=I_i\left(\frac{Z_2-Z_1}{Z_2+Z_1}\right)^2, \quad I_t=I_i\left(\frac{4 Z_2 Z_1}{Z_2+Z_1}\right)^2.
\end{equation}

To overcome the limitations of traditional methods in handling complex interfaces, we introduced the Phong model from the field of computer graphics to accurately simulate specular reflection phenomena\cite{phong1998illumination}. This model analyzes reflected light intensity by considering ambient light, diffuse reflection, and specular reflection:
\begin{equation}
\begin{aligned}
I_r & =f_r\left(\boldsymbol{I}_i, \boldsymbol{I}_o\right) I_i \\
& =k_{\mathrm{a}} I_i+k_{\mathrm{d}}(\hat{\boldsymbol{L}} \cdot \widehat{\mathbf{n}}) I_i+k_{\mathrm{s}}(\widehat{\boldsymbol{R}} \cdot \widehat{\boldsymbol{V}})^n I_i \\
& =I_{r a}+I_{r d}+I_{r s} .
\end{aligned}
\end{equation}

The Phong model's Bidirectional Reflectance Distribution Function (BRDF) exhibits rotational symmetry with respect to the reflection angle, thus satisfying $f\left(\boldsymbol{I}_i, \boldsymbol{I}_o\right)=p\left(\boldsymbol{I}_r \cdot \boldsymbol{I}_o\right)$. When ignoring complex optical phenomena such as interreflections and self-occlusions, the radiance in the view direction can be determined solely as a function of the specular reflection direction $\hat{\boldsymbol{I}}_{r s}$:
\begin{equation}
L_{\text {out }}\left(\hat{\boldsymbol{I}}_o\right) \propto \int L_{\text {in }}\left(\hat{\boldsymbol{I}}_i\right) p\left(\hat{\boldsymbol{I}}_{r s} \cdot \hat{\boldsymbol{I}}_i\right) d \hat{\boldsymbol{I}}_i=F\left(\hat{\boldsymbol{I}}_{r s}\right)
\end{equation}

Therefore, we parameterize the relationship between the specular reflection direction $\hat{\boldsymbol{I}}_{r s}$ and the viewing direction $\hat{\boldsymbol{I}}_{o}$:
\begin{equation}
\hat{\boldsymbol{I}}_{r s}=2\left(\hat{\boldsymbol{I}}_o \cdot \widehat{\mathbf{n}}\right) \widehat{\mathbf{n}}-\hat{\boldsymbol{I}}_o
\end{equation}

Where $\hat{\boldsymbol{I}}_o=-\hat{\mathbf{d}}$ represents the unit vector from the reflection point on the boundary to the transducer, and $\widehat{\mathbf{n}}$ is the normal vector at that point. In practice, the medium interface is usually not a perfect mirror but consists of many microfacets. This results in the macroscale normal vector $\widehat{\mathbf{n}}$ being a perturbed normal vector $\widehat{\mathbf{n}}^{\prime}$ , also known as the microfacet normal. As shown in Figure 2(b), the corrected normal vector $\widehat{\mathbf{n}}^{\prime}$ is calculated as follows:
\begin{equation}
\widehat{\mathbf{n}}^{\prime}=\widehat{\mathbf{n}}+\delta\left(\hat{\boldsymbol{I}}_o-\widehat{\mathbf{n}}\left(\widehat{\mathbf{n}} \cdot \hat{\boldsymbol{I}}_o\right)\right)
\end{equation}

Where $\delta$ represents the roughness at that location. By accounting for the microfacet structure of the medium, a more accurate specular reflection direction can be calculated:
\begin{equation}
\hat{\boldsymbol{I}}_{r s}=2\left(\hat{\mathbf{n}}^{\prime} \cdot \hat{\boldsymbol{I}}_o\right) \hat{\mathbf{n}}^{\prime}-\hat{\boldsymbol{I}}_o .
\end{equation}

By parameterizing the specular reflection direction $\hat{\boldsymbol{I}}_{r s}$, we input it into a multilayer perceptron (MLP), training the model to output the integrated BRDF as a function of $\hat{\boldsymbol{I}}_{r s}$. This approach not only captures the reflection characteristics of ultrasound waves on various surfaces and materials but also allows for accurate simulation of complex scenes through model training. We also supplement the output of diffuse reflection using a spatial MLP, providing a more comprehensive description of the interaction between sound waves and interfaces. To better adapt to the actual conditions of ultrasound propagation, we introduce the angle between the normal vector and the outgoing direction $\widehat{\mathbf{n}}^{\prime} \cdot \hat{\boldsymbol{I}}_o$ as an additional input parameter, enabling the model to dynamically adjust reflection characteristics.

\subsection{Reflective Harmonic Encoding}
In the traditional NeRF framework, positional encoding methods have limitations when handling high-frequency information, leading to an inability to accurately reproduce high-frequency details in the propagation of ultrasound waves\cite{huang2023refsr}\cite{mildenhall2021nerf}. This poses challenges in simulating the behavior of ultrasound waves in various media, especially in biological tissues with complex structures and heterogeneous properties, where traditional methods fail to accurately capture reflection direction information. To address this issue, we drew inspiration from the Integrated Directional Encoding (IDE) method in Mip-NeRF\cite{barron2021mip}and applied it to ultrasound imaging, calling it Reflective Harmonic Encoding (RHE).
RHE uses spherical harmonics to encode high-frequency information of reflection directions, making it particularly suitable for biological tissues with complex characteristics\cite{ahrens2012hrtf}. It finely encodes reflection directions, improving reconstruction accuracy. Specifically, we use the von Mises-Fisher (vMF) distribution to define the distribution of reflection directions and encode these directions using a set of spherical harmonics. First, we define the distribution of reflection directions on the unit sphere using the von Mises-Fisher distribution, centered at the reflection vector $\hat{\boldsymbol{I}}_r$ , with the concentration parameter $\kappa$ defined as the reciprocal of roughness $\rho$, $\kappa=1 / \rho$. Next, we represent the encoding of reflection directions using a weighted sum of spherical harmonic functions $\left\{Y_{\ell}^m\right\}$. The encoding form under the vMF distribution is defined as:
\begin{equation}
\operatorname{RHE}\left(\hat{\boldsymbol{I}}_r, \kappa\right)=\sum_{(\ell, m) \in \mathcal{L}} A_{\ell}(\kappa) Y_{\ell}^m\left(\hat{\boldsymbol{I}}_r\right)
\end{equation}

Where $\mathcal{L}$ represents the set of spherical harmonic functions. The weighting function $A_{\ell}(\kappa)$ depends on the concentration parameter $\kappa$ and can be approximated by the following formula:
\begin{equation}
A_{\ell}(\kappa) \approx \exp \left(-\frac{\ell(\ell+1)}{2 \kappa}\right) \text {. }
\end{equation}
This encoding method allows us to more accurately capture and express information about ultrasound wave reflections. Through RHE, we can finely encode these reflection directions, significantly improving the accuracy of boundary reconstructions. Especially when dealing with reflections in complex tissue structures and rich details, RHE provides higher resolution and accuracy. 

\subsection{Sine Activation Function}
Sinusoidal representation networks (SIRENs)\cite{sitzmann2020implicit} utilize sine as the activation function in MLPs to model high-frequency information accurately. They introduced an initialization scheme to prevent vanishing gradients in periodic activation functions, showing that SIRENs fit complex signals and their derivatives robustly. This provides a solid foundation for using the Sine activation function in ultrasound image reconstruction.The periodicity and continuity of the Sine activation function enable it to capture high-frequency components and subtle variations effectively. Its expression is:
\begin{equation}
\text { Sine }(x)=\sin (x)
\end{equation}

Compared to traditional activation functions like ReLU and Sigmoid, which are sensitive to initial parameters and unstable with complex signals, the Sine activation function is more stable and robust over a wide range of parameters\cite{parascandolo2016taming}.
In the Ultra-NeRF model, results were initially unstable and highly dependent on network configuration. By introducing the Sine activation function, our model demonstrated more consistent performance across different initial configurations, reducing the impact of initial settings on the final imaging outcome. This improvement enhances the model's robustness and stability, making it better suited for various configurations and complex ultrasound imaging scenarios.

\subsection{Ultrasound Neural Rendering Architecture}
To address the complexity of ultrasound scenes and accurately reconstruct ultrasound characteristics within the medium, we designed an ultrasound neural rendering architecture based on implicit neural networks and volume rendering (as shown in Figure 3). This architecture integrates two main modules, the directional MLP and the spatial MLP, and performs volume rendering based on ray tracing and physical principles, significantly improving the reconstruction accuracy of ultrasound images.

The primary function of the directional MLP is to process ray direction information. Specifically, it receives the ray direction $\hat{\mathbf{d}}$ as input and generates view-dependent reflection intensity information. By generating the reflection intensity estimate $\widehat{\boldsymbol{C}}_{r e f}$, it ensures that the reflection characteristics under each view are accurately captured, enhancing the ability to simulate complex reflection properties. The formula is as follows:
\begin{equation}
\widehat{\boldsymbol{C}}_{r e f}=\mu\left(\mathbf{c}_d+\mathbf{c}_s\right)
\end{equation}

Where $\widehat{\boldsymbol{C}}_{r e f}$ is the reflection intensity estimate, $\mathbf{c}_s$ is the specular reflection information encoded by direction, $\mathbf{c}_d$ is the diffuse reflection information produced during the interaction of ultrasound waves with the interface, which is only position-dependent, and $\mu$ is a scaling function that limits the output value to the range [0,1].

The spatial MLP, on the other hand, is responsible for predicting a set of physical property parameters, including the attenuation coefficient $\alpha$, reflectivity $\beta$, scattering density $\rho_s$, and amplitude $\phi$. We apply Fourier feature positional encoding to map the spatial position $\mathbf{x}$ into a high-dimensional space\cite{tancik2020fourier}, capturing high-frequency details in ultrasound propagation. This information is then fed into the spatial MLP to generate the aforementioned physical property parameters, accurately reconstructing the ultrasound characteristics of the medium. During the rendering process, these physical properties are used to simulate the propagation and reflection behavior of ultrasound waves in different media, resulting in the rendering outcome $\widehat{\boldsymbol{C}}_{\text {render }}$. Finally, the reflection information generated by the directional MLP is combined with the rendering outcome to produce the final ultrasound imaging output.
\begin{figure*}[t]
\centerline{\includegraphics[width=\columnwidth]{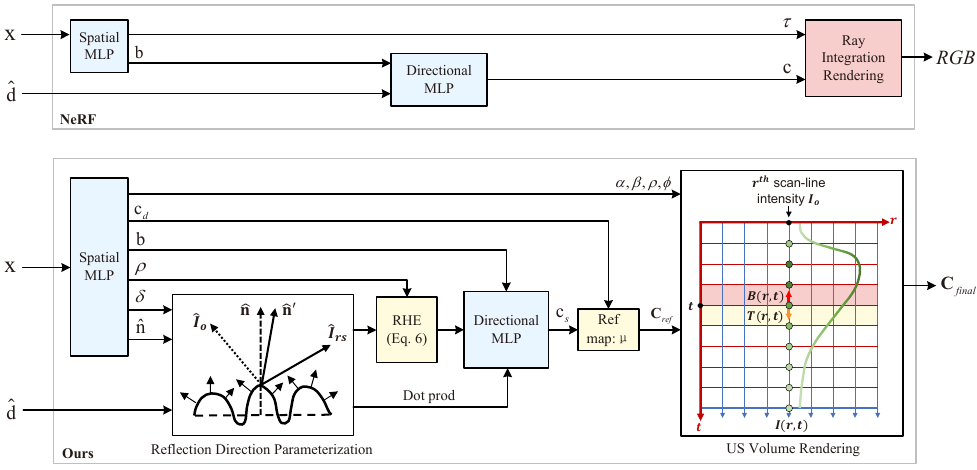}}
\caption{Ultrasound Neural Rendering Architecture of UlRe-NeRF. This figure illustrates our proposed ultrasound neural rendering architecture. The view-dependent reflection intensity information and physical property parameters generated by the directional MLP and spatial MLP modules are combined in the rendering section, achieving high-fidelity simulation of ultrasound wave propagation and reflection behavior.}
\label{fig3}
\end{figure*}
By improving the network architecture and rendering process, UlRe-NeRF not only accurately captures and simulates the propagation and reflection behavior of ultrasound waves in different media but also dynamically adapts to view changes, enhancing the detail and accuracy of the reconstructed images. This approach excels in handling complex ultrasound imaging effects and enhances the model's robustness and adaptability, achieving high-fidelity simulation of ultrasound wave propagation and reflection.

\subsection{US volume rendering}
The traditional Neural Radiance Fields (NeRF) method generates images by accumulating point data in 3D space along the line of sight. However, this approach has significant limitations when applied to ultrasound imaging. The physical phenomena involved in ultrasound imaging are more complex, requiring consideration of wave effects and acoustic impedance matching, which are not factors in natural image rendering. Therefore, a more specialized approach is necessary. We improved the Ultra-NeRF method by proposing a volume rendering technique that utilizes the physical properties of the medium. This new method combines parameters generated by directional MLP and spatial MLP and incorporates classical ultrasound imaging theory and ray tracing technology\cite{burger2012real}, thereby achieving more accurate simulations of ultrasound wave propagation and reflection to produce high-fidelity rendered images.

 Specifically, we synthesize views by querying the ultrasound characteristics of the medium along the path of the sound wave: attenuation coefficient $\alpha$, reflectivity $\beta$, scattering density $\rho_s$, and amplitude $\phi$. As shown in Figure 3, we treat ultrasound waves as rays emitted from the transducer, simulating their propagation paths through different media. For each scanline $r$ , we define the ultrasound intensity $\boldsymbol{C}_{\text {render }}(r, t)$ at a distance $t$ as the sum of the reflected energy $R(r, t)$ and backscattered energy $B(r, t)$, describing the energy changes of the ultrasound wave during propagation:
\begin{equation}
\boldsymbol{C}_{\text {render }}(r, t)=R(r, t)+B(r, t)
\end{equation}

Based on the Beer-Lambert law, we introduce the concept of residual energy $I(r, t)$, whose variation is calculated using the following formula:
\begin{equation}
I(r, t)=I_0 \cdot \exp \left(-\int_0^{t-1}(\alpha \cdot f \cdot \Delta t) d \tau\right) \cdot \prod_{n=0}^{t-1}(1-\beta(r, n))
\end{equation}

Where $\alpha$ is the attenuation coefficient of the medium, $f$ is the frequency of the ultrasound wave, and $\beta(r, n)$ is the reflection coefficient of the medium in the tissue. This formula describes the exponential decay of the signal as it propagates through the medium.

The reflected energy $R(r, t)$ refers to the energy generated at the medium interface due to reflection. The calculation of $R(r, t)$ is based on the residual energy $\beta(r, t)$, the reflection coefficient , and a predefined two-dimensional point spread function (PSF):
\begin{equation}
R(r, t)=I(r, t) \cdot \beta(r, t) \otimes \operatorname{PSF}(r),
\end{equation}

In our model, the point spread function (PSF) is simplified to include only the axial and lateral directions, excluding the elevation direction. The PSF function is approximated as a two-dimensional Gaussian function modulated by a cosine function, specifically:
\begin{equation}
\operatorname{PSF}(x, y)=\exp \left(-\frac{1}{2}\left(\frac{x^2}{\sigma_x^2}+\frac{y^2}{\sigma_y^2}\right)\right) \cdot \cos (2 \pi f x)
\end{equation}

Where $\sigma_x$ and $\sigma_y$ represent the standard deviations in the axial and lateral directions, respectively, and $f$ is the frequency of the ultrasound wave.

Backscattered energy  $B(r, t)$ refers to the energy generated by scattering, which consists of the residual energy  $I(r, t)$ and the two-dimensional distribution of scatter points  $T(r, t)$:
\begin{equation}
B(r, t)=I(r, t) \cdot S(r, t) \otimes \operatorname{PSF}(r),
\end{equation}

Here, the two-dimensional distribution of scatter points $S(r, t)$ is defined as:
\begin{equation}
S(r, t)=\eta(r, t) \cdot \phi(r, t)
\end{equation}

where the function $\eta(r, t)$ indicates the intensity of the scatterers. To add a degree of smoothness while maintaining boundary clarity, we use the Beta distribution to parameterize the scattering density $\rho_s$, which provides better continuity compared to the Bernoulli distribution. The intensity of the scatter points is determined by their amplitude $\phi$.

By integrating these energy estimates, we obtain the ultrasound intensity at different positions along each scanline. By normalizing and combining the intensity values $\widehat{\boldsymbol{C}}_{\text {render }}$ generated by the rendering process with the reflection intensity $\widehat{\boldsymbol{C}}_{\text {ref }}$ produced by the directional MLP, we get the final rendered result:
\begin{equation}
\widehat{\boldsymbol{C}}_{\text {final }}=\mathcal{G}\left(\widehat{\boldsymbol{C}}_{\text {render }}+\widehat{\boldsymbol{C}}_{\text {ref }}\right) .
\end{equation}

This improved method not only addresses the shortcomings of Ultra-NeRF in handling complex artifacts and reflection characteristics but also enhances rendering consistency and accuracy across different views and complex scenes. Through this new approach, UlRe-NeRF can more accurately simulate the propagation and reflection behavior of ultrasound waves in various media, significantly improving the physical realism and visual quality of the images.

\section{Experiments}
\subsection{Experimental Setup}
Our experiments were conducted on a computer equipped with an NVIDIA RTX 3070 GPU and 32GB of RAM. We utilized the Ultra-NeRF open-source dataset to simulate liver ultrasound images generated from CT scans, following the image generation methods outlined in Ultra-NeRF's work and using the ImFusion tool. The dataset includes seven scans: six at different oblique angles and one vertical scan. Each scan contains 200 2D ultrasound images along with their tracking information. The oblique scans provide images from different perspectives and directions, with frames containing occlusions caused by scanning between ribs. We used four oblique scans (800 frames) for training and one vertical scan plus two oblique scans (600 frames in total) for testing.
We employed the same quantitative metrics as previous view synthesis studies\cite{barron2021mip}\cite{mildenhall2021nerf}\cite{zhang2021physg}. Performance evaluation metrics include Peak Signal-to-Noise Ratio (PSNR), Structural Similarity Index (SSIM)\cite{wang2004image}, Learned Perceptual Image Patch Similarity (LPIPS)\cite{zhang2018unreasonable}, and Mean Squared Error (MSE). Among these, LPIPS focuses on the global structure and noise levels of the images and performs exceptionally well in measuring the similarity of images with significant differences and noticeable disturbances, being highly correlated with human perceptual preferences\cite{mier2021deep}.

\subsection{Ablation Study Results of the Neural Network Architecture}
Although NeRF and Mip-NeRF (the previous best-performing techniques) have shown outstanding performance in view synthesis and image reconstruction, their ray integration-based rendering method has limitations in ultrasound imaging. The traditional NeRF method determines pixel color values by integrating along a ray, which is suitable for visible light images. However, ultrasound imaging involves interactions of ultrasound waves with different tissues along the propagation path. To address this, we made specific modifications to NeRF and Mip-NeRF to better handle the characteristics of ultrasound images. Specifically, we used a Directional MLP in the model to directly output simulated ultrasound images, capturing reflection information of ultrasound waves in different directions. Furthermore, we conducted ablation studies on the performance of each model using the ultrasound dataset.

\begin{figure*}[t]
\centerline{\includegraphics[width=\columnwidth]{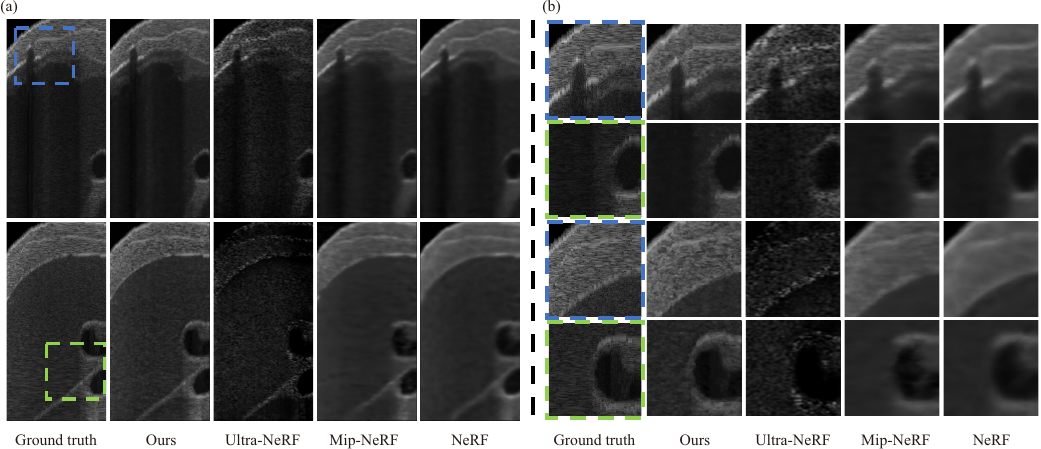}}
\caption{Our model demonstrates accurate and high-fidelity performance, which highlights two different frames of ultrasound images. (a) Overall image comparison; (b) Detailed section comparison.}
\label{fig4}
\end{figure*}

In Figure 4, we compare the performance of different methods in ultrasound imaging. As shown in Figure 4(a), NeRF fails to accurately capture the complex reflection and scattering phenomena in ultrasound images, resulting in blurred image edges. Although Mip-NeRF shows some improvement, it still lacks refinement, with noticeable high-frequency detail and fine feature loss. Ultra-NeRF improves in detail restoration and visual effect but still suffers from noise and blurriness when handling complex tissues, and its robustness is poor, with performance heavily dependent on model initialization. Figure 4(b) shows a detailed comparison, where only our model accurately captures reflection and scattering information, exhibiting higher image clarity and less noise. By introducing the reflection direction parameterization and RHE, our method excels in restoring high-frequency details and edge information, with less noise and overall visual effects closer to real scenes, achieving high-fidelity image reconstruction. This advantage is particularly evident in the LPIPS metric.
\begin{table}[ht]
\centering
\caption{Baseline Comparisons and Ablation Study}
\label{tab1}
\begin{tabular}{l|ccc|c}
\multicolumn{1}{c|}{}  & LPIPS↓                        & SSIM↑                         & PSNR↑                         & MSE↓                            \\ \hline
NeRF                   & 0.565                         & \cellcolor[HTML]{FFFFB2}0.570 & \cellcolor[HTML]{FFFFB2}29.66 & \cellcolor[HTML]{FFFFB2}1.43e-3 \\
Mip-NeRF               & 0.529                         & 0.561                         & 29.41                         & 1.49e-3                         \\
Ultra-NeRF             & \cellcolor[HTML]{FFFFB2}0.321 & 0.449                         & 20.51                         & 5.85e-2                         \\ \hline
Ours, no reflection    & 0.553                         & 0.486                         & 22.84                         & 8.57e-3                         \\
Ours, concat. viewdir  & 0.566                         & 0.480                         & 22.40                         & 1.01e-2                         \\
Ours, dotprod          & 0.551                         & 0.493                         & 23.10                         & 8.30e-3                         \\
Ours, no roughness     & \cellcolor[HTML]{FFD9B2}0.261 & 0.544                         & 28.71                         & 1.78e-3                         \\
Ours, no diffuse color & 0.552                         & 0.494                         & 23.13                         & 8.26e-3                         \\ \hline
Ours, ReLu             & 0.547                         & \cellcolor[HTML]{FFD9B2}0.585 & \cellcolor[HTML]{FFD9B2}29.67 & \cellcolor[HTML]{FFD9B2}1.41e-3 \\
Ours, PE               & 0.553                         & 0.494                         & 23.12                         & 8.29e-3                         \\
Ours, ReLu. PE         & 0.552                         & \cellcolor[HTML]{FFB2B2}0.586 & \cellcolor[HTML]{FFB2B2}29.71 & \cellcolor[HTML]{FFB2B2}1.40e-3 \\
Ours                   & \cellcolor[HTML]{FFB2B2}0.250 & 0.546                         & 28.75                         & 1.78e-3                        
\end{tabular}
\end{table}
The quantitative results in Table 1 indicate that our model significantly outperforms others in the LPIPS metric, demonstrating a substantial advantage in perceptual image quality. The various improvement modules described in section 3 each contribute to this performance. When using view direction instead of reflection direction as input to the Directional MLP ("no reflection"), the model's reconstruction metrics notably decrease, highlighting the benefit of parameterizing reflective radiance. Adding view direction as an additional input to the Directional MLP ("concat. viewdir") severely impacts performance, underscoring the importance of parameterizing specular appearance as a function of view direction. Moreover, not feeding $\widehat{\mathbf{n}}^{\prime} \cdot \hat{\boldsymbol{I}}_o$ into the Directional MLP leads to significant degradation in rendering. Using fixed RHE weight parameters slightly reduces performance, demonstrating the robustness of spherical harmonic encoding. Removing diffuse color ("no diffuse color") greatly affects performance, further proving the crucial role of diffuse reflection in a complete reflection model. Finally, replacing the Sine activation function with the ReLU activation function used in NeRF also impacts performance. Deleting RHE structural components (roughness, diffuse color, etc.) and replacing them with NeRF's standard positional encoding ("PE") shows a performance drop, further validating the effectiveness of our proposed improvements. The combination of ReLU activation function and positional encoding ("ReLu. PE") also shows a decline in performance. Our improvements significantly enhance detail restoration and overall image quality by better capturing high-frequency details and accurately representing ultrasound wave reflection information. Qualitative and quantitative analyses clearly show the superiority of our model in ultrasound image reconstruction tasks.

\subsection{Ablation Study Results of the Rendering Components}

\begin{table}[t]
\centering
\caption{Rendering Method Comparisons and Ablation Study}
\label{tab2}
\begin{tabular}{l|lll|l}
                             & LPIPS↓                        & SSIM↑                         & PSNR↑                         & MSE↓                            \\ \hline
Ultra-NeRF                   & 0.321                         & 0.449                         & 20.51                         & 5.85e-2                         \\
Ultra-NeRF, our rendering    & \cellcolor[HTML]{FFD9B2}0.314 & \cellcolor[HTML]{FFD9B2}0.498 & 22.31                         & 4.75e-2                         \\
Ours, Ultra-NeRF’s rendering & 0.361                         & 0.409                         & \cellcolor[HTML]{FFD9B2}24.97 & \cellcolor[HTML]{FFD9B2}4.20e-3 \\
Ours                         & \cellcolor[HTML]{FFB2B2}0.250 & \cellcolor[HTML]{FFB2B2}0.546 & \cellcolor[HTML]{FFB2B2}28.75 & \cellcolor[HTML]{FFB2B2}1.78e-3
\end{tabular}
\end{table}

To demonstrate the effectiveness of UlRe-NeRF's rendering method in ultrasound image reconstruction, we designed various ablation experiments comparing it to Ultra-NeRF. These experiments included replacing Ultra-NeRF's rendering method with ours and using Ultra-NeRF's rendering method in our model. Table 2 shows the quantitative comparison results of different models across several metrics. It is evident that UlRe-NeRF exhibits significant advantages in all metrics, especially in significantly reducing the LPIPS value. This improvement is reflected in the visual quality of the images, with a noticeable suppression of noise and artifacts, resulting in clearer and more realistic images. In contrast, using Ultra-NeRF's original rendering method results in significant noise and detail loss in the images. Additionally, when Ultra-NeRF adopts our method, there is a significant improvement in image quality and metrics, indicating that our method better reconstructs ultrasound signals and effectively reduces noise interference. Quantitative analysis of the results shows that our method can more accurately capture and reconstruct details when handling complex tissue structures, significantly enhancing image clarity and realism.

\begin{figure}[!t]
\centerline{\includegraphics[width=\columnwidth]{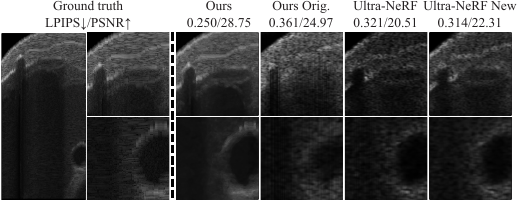}}
\caption{Visual Comparison of Different Rendering Methods. Ours Orig. indicates the use of Ultra-NeRF's rendering method, while Ultra-NeRF New indicates the use of our rendering method.}
\label{fig5}
\end{figure}

\begin{figure}[t]
\centerline{\includegraphics[width=\columnwidth]{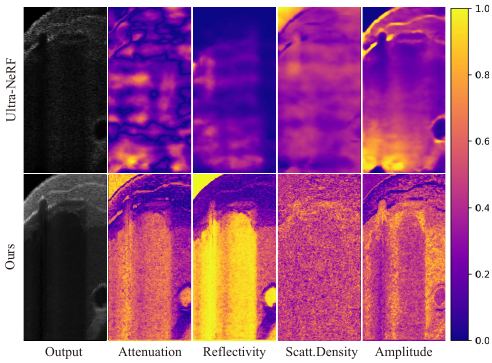}}
\caption{Optimization Effects on Various Physical Property Parameters. We used normalized values to unify the scale for display in the same dimension.}
\label{fig6}
\end{figure}

Figure 5 presents the visual effects of four different rendering methods in ultrasound images. Images generated by Ultra-NeRF exhibit high levels of noise and shadow issues, particularly with blurred edges and details, leading to overall lower image quality. When Ultra-NeRF's rendering method is replaced with ours, there is a reduction in shadows and an improvement in detail clarity, although some noise remains. In contrast, our volume rendering method performs best in detail restoration and noise reduction, significantly enhancing the overall image quality. This improvement in visual effects is not only reflected in the clarity of details and edges but also in the overall contrast and visual consistency of the images. Using our rendering method, ultrasound images can better preserve artifacts and real details when dealing with complex tissue structures, resulting in high-fidelity reconstructed images.

Figure 6 further illustrates the optimization effects on various physical property parameters, including attenuation coefficient $\alpha$, reflectivity $\beta$, scattering density $\rho_s$, and amplitude $\phi$. For the attenuation coefficient $\alpha$, our method provides a more accurate description of attenuation behavior, thereby improving image contrast and clarity. The optimized reflectivity $\beta$ significantly enhances the precision of interface reflections, making reflection boundaries sharper and more distinct, ensuring clearer delineation between tissues. The scattering density $\rho_s$ optimized using the beta distribution more effectively describes scattering phenomena compared to the Bernoulli distribution, reducing blurriness caused by scattering. Finally, the optimized amplitude $\phi$ parameter improves signal consistency, resulting in clearer details within the image, especially in areas with complex tissue structures, further enhancing the realism and visual quality of the images.

\section{Discussion}

In this paper, we present UlRe-NeRF, a more effective and precise Neural Rendering method for 3D Ultrasound Imaging. We incorporate additional inductive biases about ultrasound: enhancing the implicit representation network with reflection direction parameterization and Reflective Harmonic Encoding (RHE) to capture complex tissue reflections, and improving explicit volumetric rendering by integrating reflection intensity and volume rendering to simulate the medium’s physical properties and light interactions.

Our experiments show that UlRe-NeRF achieves high-fidelity 3D ultrasound imaging synthesis. It effectively learns view-dependent geometry, captures complex tissue structures, simulates various physical properties of the medium, and accurately reproduces the propagation and reflection behavior of ultrasound waves. This study marks an important advance in ultrasound image reconstruction techniques, demonstrating the potential of ultrasound neural rendering architectures based on implicit neural networks and explicit ultrasound volume rendering to improve image quality.

However, the inherent characteristics of ultrasonic imaging limit the usage of hierarchical sampling strategies. Thus, methods specifically designed for ultrasound rendering to enhance sample efficiency are required. Although we have introduced new explicit ultrasound volume rendering techniques to improve contrast and clarity in physical property simulations, further advancements are needed to develop more efficient and precise methods for simulating the propagation and reflection behavior of ultrasound waves. Additionally, in more complex freehand scanning situations, where ultrasound data is sparse, additional regularization for view synthesis from sparse inputs is necessary.

\section{Conclusion}
The UlRe-NeRF model represents a significant advancement in 3D ultrasound imaging by combining implicit neural networks with explicit volumetric rendering. Our approach successfully addresses the limitations of traditional ultrasound imaging methods, providing high-fidelity reconstructions of complex tissue structures and realistic simulations of wave propagation. While our work has made considerable progress, further research is needed to improve sample efficiency and address challenges associated with sparse data in freehand scanning scenarios. We anticipate that this study will inspire continued exploration and refinement of neural rendering techniques for medical applications, contributing to the future development of ultrasound imaging technology.



\bibliographystyle{splncs04}
\bibliography{reference}
%




\end{document}